\documentclass{article}

\usepackage{PRIMEarxiv}

\usepackage[utf8]{inputenc} % allow utf-8 input
\usepackage[T1]{fontenc}    % use 8-bit T1 fonts
\usepackage{hyperref}       % hyperlinks
\usepackage{url}            % simple URL typesetting
\usepackage{booktabs}       % professional-quality tables
\usepackage{amsfonts}       % blackboard math symbols
\usepackage{nicefrac}       % compact symbols for 1/2, etc.
\usepackage{microtype}      % microtypography
\usepackage{lipsum}
\usepackage{fancyhdr}       % header
\usepackage{graphicx}       % graphics
\graphicspath{{images/}}     % organize your images and other figures under media/ folder
\usepackage{subcaption}
\usepackage{tabularray}
\usepackage{graphicx} % Required for including images
\usepackage{booktabs} % For prettier tables
\usepackage{algorithm}
\usepackage{algpseudocode}
\usepackage{array}
\title{ASGM-KG: Unveiling Alluvial Gold Mining Through Knowledge Graphs
}

\author{
  Debashis Gupta, Aditi Golder, Luis Fernendez, Miles Silman, \\Greg Lersen, Fan Yang, Bob Plemmons, Sarra Alqahtani, Paul Victor Pauca \\
  Wake Forest University \\
  Winston-Salem, NC \\
  \texttt{\{guptd23, golda24, fernanle, silmanmr, larseng, yangfan, plemmons, sarra-alqahtani, paucavp\}@wfu.edu} \\
  \And
  Sakib Imtiaz \\
  Rajshahi University of Engineering and Technology \\
  Rajshahi, Bangladesh \\
  \texttt{sakibimtiaz1998@gmail.com} \\
}

\begin{document}
\maketitle

\begin{abstract}

Artisanal and Small-Scale Gold Mining (ASGM) is a low-cost yet highly destructive mining practice, leading to environmental disasters across the world's tropical watersheds. The topic of ASGM spans multiple domains of research and information, including natural and social systems, and knowledge is often atomized across a diversity of media and documents. We therefore introduce a knowledge graph (ASGM-KG) that consolidates and provides crucial information about ASGM practices and their environmental effects. 
The current version of ASGM-KG consists of 1,899 triples extracted using a large language model (LLM) from documents and reports published by both non-governmental and governmental organizations. These documents were carefully selected by a group of tropical ecologists with expertise in ASGM. This knowledge graph was validated using two methods. First, a small team of ASGM experts reviewed and labeled  triples as factual or non-factual. Second, we devised and applied an automated factual reduction framework that relies on a search engine and an LLM for labeling triples. Our  framework performs as well as five baselines on a publicly available knowledge graph and achieves over $90\%$ accuracy on our ASGM-KG validated by domain experts. ASGM-KG demonstrates an advancement in knowledge aggregation and representation for complex, interdisciplinary environmental crises such as ASGM.

\end{abstract}

% keywords can be removed
% \keywords{First keyword \and Second keyword \and More}

\section{Introduction}

\begin{figure}[htbp]
\includegraphics[width=\linewidth]{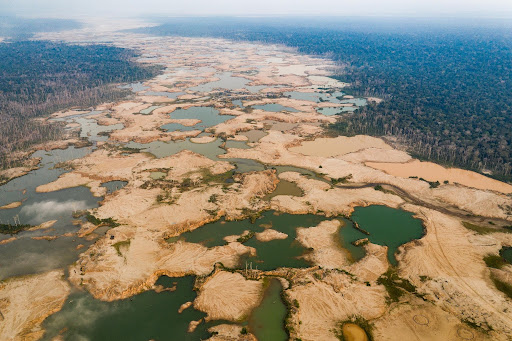}
\caption{Aerial view of La Pampa, an ASGM hot zone in Madre de Dios, Per\'u. (Photo by Jorge Caballero) }
\label{fig:asgm_pic}
\end{figure}

Artisanal and small-scale gold mining (ASGM) and the global demand for gold constitute an 
emerging threat to the conservation and preservation of tropical systems worldwide~\cite{dethier2019heightened,alvarez2015global}. 
ASGM is a rudimentary mining practice that extracts   gold particles from alluvial  sediments. %It is generally associated with land cover/land use (LCLU) change that can encompass large areas, including the clearing of primary tropical rainforest.
This involves removing all above-ground vegetation and soil, using large quantities of water and low-tech equipment, such as suction pumps, to wash large quantities of sediment down a sluice, and finally employing mercury to amalgamate the gold particles. %At the end of the operation, the amalgam is heated to vaporize most of the mercury. 
All remaining mercury-contaminated material are   vaporized to the air or released to soil and water  before moving the operation to another site.
%It requires the removal of all biomass above ground, and the use of water to wash large amounts of sediment down a sluice, where the heavier gold particles are trapped by a carpet material. The carpet material is removed by the end of the operation and manually submerged in a large solution of mercury and water to amalgamate the gold. Finally, the amalgam is extracted and heated to vaporize the mercury. The remaining material is released on the ground before moving the operation to another zone.
Shallow  water tables quickly fill any excavations, leaving behind a mixture of ponds and bare earth~\cite{gerson2020artificial} (See Figure~\ref{fig:asgm_pic}). 

\begin{table*}[tbp]
\caption{ Data Sources}
\label{tab:sources}
\setlength{\tabcolsep}{5pt}
\resizebox{\textwidth}{!}{%
\begin{tabular}{@{}lclc@{}} % Removed one "c" from the column format specifier
\toprule
\multicolumn{1}{c}{\textbf{Source}} & \textbf{Language} & \textbf{Type} & \multicolumn{1}{c}{\textbf{Link}} \\
\midrule
WWF Report & English & pdf & \url{https://wwfint.awsassets.panda.org/downloads/healthy_rivers_healthy_people.pdf} \\
UN Environment & English & pdf & \url{https://wedocs.unep.org/bitstream/handle/20.500.11822/29831/gma_tech.pdf?sequence=1&isAllowed=y} \\
WWF Report & English & pdf & \url{https://d2ouvy59p0dg6k.cloudfront.net/downloads/reporte_eng.pdf} \\
USAID & English & pdf & \url{https://usaidgems.org/Documents/SectorGuidelines/Mining_Guidelines_20170630_Final.pdf} \\
University of California Press & English & pdf & \url{https://online.ucpress.edu/elementa/article/doi/10.1525/elementa.274/112794/Senegalese-artisanal-gold-mining-leads-to-elevated} \\
NRDC & English & pdf & \url{https://www.nrdc.org/sites/default/files/investing-artisanal-gold-summary.pdf} \\
UN Environment & English & pdf & \url{https://wedocs.unep.org/bitstream/handle/20.500.11822/27579/GMA2018.pdf?sequence=1&isAllowed=y} \\
World Gold Council & Spanish & pdf & \url{https://www.gold.org/esg/artisanal-and-small-scale-gold-mining} \\
USP NEWSPAPER & Portuguese & website & \url{https://jornal.usp.br/ciencias/mercurio-do-garimpo-causa-danos-neurologicos-aos-yanomami/}\\
\bottomrule
\end{tabular}
}
\end{table*}

Remote sensing data, such as airborne and satellite imagery, have been primary data sources for investigating the scale and effects of ASGM~\cite{caballero2018deforestation,dethier2019heightened,dethier2023global,kimijima2021detection}.
Studies of historical satellite imagery have shown that the pace of ASGM activity has increased significantly over the past decade. Landsat data products have demonstrated that ASGM was responsible for the removal of over $120,000$ hectares of primary tropical forest (approximately $1.5$ times the size of New York City) in the Madre de Dios department of Peru, from 1984 to 2017 ~\cite{caballero2018deforestation}.
Similar satellite data also shows  ASGM expanding in many other countries in South America~\cite{bruno2020artisanal,quash2024assessing}, Africa~\cite{owusu2018spatial}, and Southeast Asia~\cite{bounliyong2021k,kimijima2021detection}.
%According to the UN Environment Programme~\cite{unep2019gma2018}, each year over 2000 tons of mercury to air, water, and land.
Other remote sensing studies have found its impact on biodiversity, water quality and human health to be profound at a global scale~\cite{dethier2023global,gonzalez2019mercury,soe2022mercury}. 

Many important questions involving smaller-scale changes, such as state of forest recovery, repeated mining over previously worked sites, and forecasting future mining activity remain difficult to address via remote sensing  alone. Despite the large amounts of multimodal and multiresolution data  acquired in a daily basis for land cover monitoring, the information needed to answer these questions is hard to disambiguate without the use of additional contextual information. 
Meanwhile, research findings published in scientific venues are hard for the general public to interprete from a governance and policy standpoint to yield meaningful information and robust evidence to support policy recommendations.
Thus, new approaches are needed to support the combination of disparate data and sources of relevant knowledge~\cite{gil}.

In this resource paper, we introduce ASGM-KG, a knowledge graph specifically designed to help better understand the effects of ASGM in tropical forests, such as in the Amazon Basin that includes most countries in South America.
This effort is led by a team of computer scientists, ecologists, and domain experts with decades of expertise in ASGM and extensive field work in the tropical forests of Madre de Dios, Per\'{u}, one of the world's hotspots for legal and illegal alluvial gold mining. Through the Center for Amazonian Scientific Innovation (CINCIA) and collaboration with local governmental and non-governmental organizations, the team has been able to characterize several aspects of ASGM concerning deforestation, mercury contamination, and ecosystem restoration and recovery. 

Our work in developing ASGM-KG has two main objectives. First, it aims to provide a publicly available resource for government officials, local agencies and organizations to help guide their decision processes and to design successful interventions~\cite{dethier2023operation}. Second, it seeks to build a source of accurate information  from reputable sources, validated by domain experts, to leverage advancement in knowledge-infused deep learning and neuro-symbolic computing.

\subsection{Related Work}

Knowledge graphs are increasingly being used in remote sensing for various purposes.
%They enable integration of multimodal data through a unified semantic framework. 
Deng et al.~\cite{deng2021gakg} introduced GAKG, a large-scale multimodal academic knowledge graph, to comprehensively integrate knowledge found in the geoscience literature.
WorldKG ~\cite{dsouza2021worldkg} is a new geographic knowledge graph built from the OpenStreetDataset to provide semantic representation of geographic entities. 

As is the case in alluvial gold mining, accurate prediction of future events is a key objective in many remote sensing applications. 
Forest fire prediction~\cite{chen2022knowledge}, landslide prediction~\cite{chen2023improved,xu2023landslide}, 
exploration of  iron and gold deposits~\cite{ironm1,goldm1}, and agriculture~\cite{xiaoxue2019review,zhang2021research}
are some specific areas where knowledge graphs and domain expertise are being utilized, often in conjunction with other types of data. However, resources, such as GAKG and WorldKG, and prediction techniques developed so far are not tailored specifically for the study of ASGM.

\section{The ASGM Knowledge Graph}

The ASGM-KG was constructed using 9 documents selected by a domain expert at CINCIA. These documents contain approximately 930-pages  in total with an average word count of 500 words per page. Links to these documents are provided in Table~\ref{tab:sources} and can also be found in our Github repository \footnote{\url{https://github.com/Debashis-Gupta/ASGM-KG-Unveiling-Alluvial-Gold-Mining-Through-Knowledge-Graphs}}. 

This initial version of ASGM-KG consists of 1650 unique entities and 785 unique relationships. A total of 43\% of entities and 29\% relations are newly discovered, i.e., they are not found in Wikidata~\cite{vrandevcic2014wikidata}.
There are 1899 triples \texttt{(subject, predicate, object)} stored in Resource Description Framework (RDF) format. %Figure~\ref{fig:Total-KG} gives a graphical \fanchange{illustration} of the \fanchange{developed} ASGM-KG. 
The corresponding data can be obtained, explored, and visualized via the ASGM-KG online portal~\footnote{\url{https://asgm-kg.streamlit.app/}}. Three downstream tasks are implemented:query answering, subgraph summarization, and chat via natural language.

% \begin{figure}[htbp]
%   \centering
%   \includegraphics[width=3.2in]{images/asgm_kg.png}
%   \caption{The ASGM-KG}
%   \label{fig:Total-KG}
% \end{figure}

Next, we describe the two major steps involved in the construction of ASGM-KG: 1) extraction of RDF statements using a large language model (LLM), and 2) factual reduction of the RDF statements using an unsupervised approach, Data Assessment Semantics (DAS), developed specifically for this task. 

\subsection{RDF Statement Extraction Using LLMs}

We extract RDF statements in \texttt{entity-relation-entity} format by querying an LLM using a specified ontology schema~\cite{Text2kgbench}, which instructs the LLM the steps to extract RDF statements from raw text. (See Algorithm~\ref{prompt:1}.)
\begin{algorithm}[htp] 
\caption{Extract RDF Triples}
\begin{algorithmic}[1]
\State \textbf{Input:} Unstructured Text
\State \textbf{Output:} RDF Table with columns [Subject, Predicate, Object]
% \Require Text (unstructured text)
% \Ensure RDF Table with columns [Subject, Predicate, Object]
\State Initialize RDF Table;
\For{\textit{each sentence in Unstructured Text}}
    \State Identify nouns and treat as entities;
    \State Replace pronouns with corresponding nouns as per context;
    \State Limit entities to a maximum of two words;
    \State Identify verbs and treat as relations;
    \State Verbs can be associated with prepositions;
    \State Limit relations to a maximum of two words;
    \State Predicate can have negation property;
    \State Preserve the order of entities and relations;
    \State Construct triples [Subject, Predicate, Object];
    \State Discard triples that do not meet predefined constraints;
    \State Append only valid triples to the RDF Table;
\EndFor
\State \textbf{return} RDF Table
\end{algorithmic}
\label{prompt:1}
\end{algorithm}

The application of this prompt tuning process resulted in $2,653$ RDF triples.
%, effectively transforming the extensive unstructured textual content of the documents into a structured format, conducive to further analysis and integration into the ASGM-KG. 
This approach to extracting RDF statements is becoming prevalent due to the remarkable capabilities of LLMs for natural language understanding, contextual processing, and semantic reasoning~\cite{pan2023large}.

\subsection{Factual Reduction via DAS}

\begin{figure}[htbp]
  \centering
  \includegraphics[width=\columnwidth]{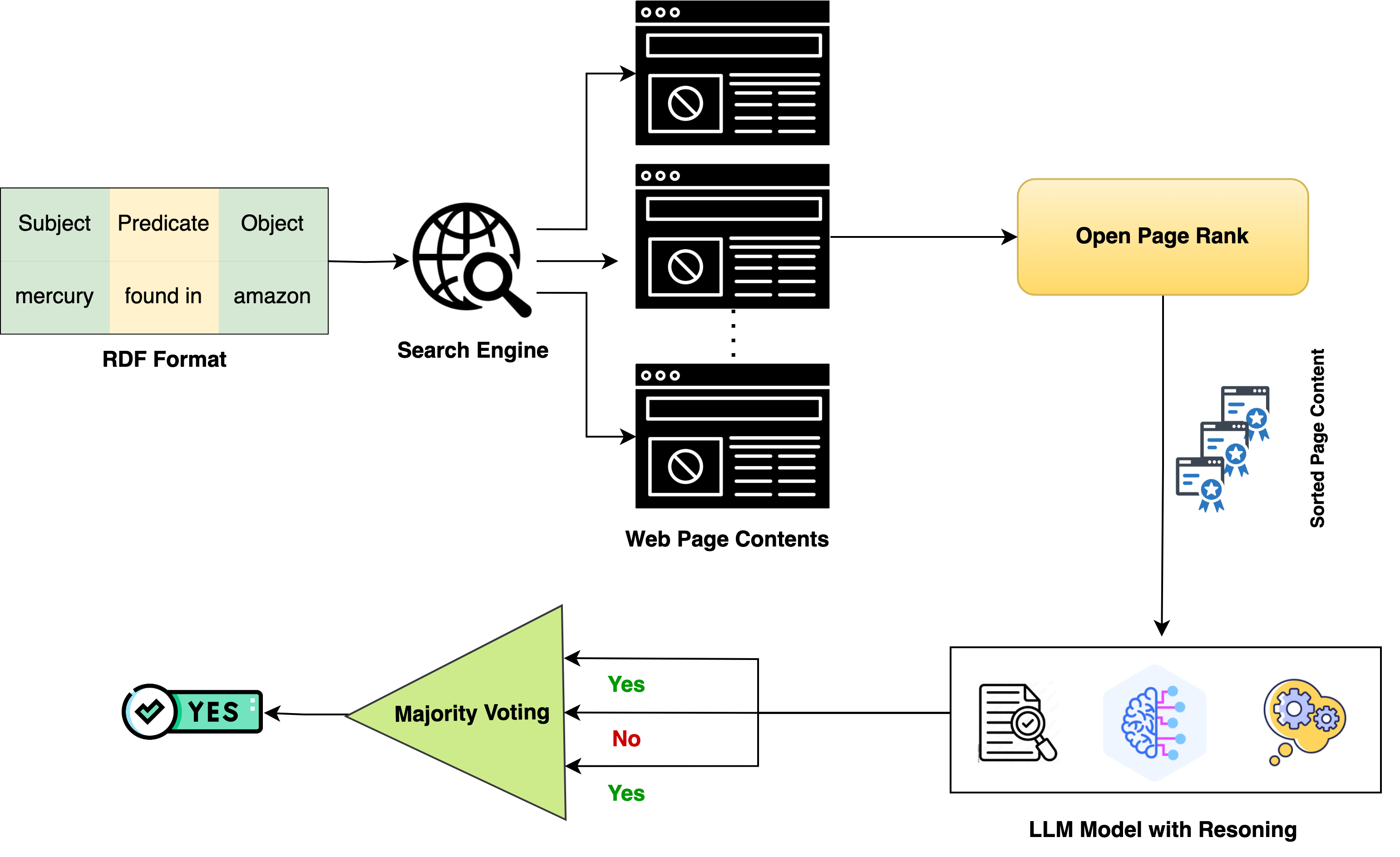}
  \caption{Data Assessment Semantics framework: Automated process for factual validation via open-source knowledge. }
  \label{fig:dasValidation}
\end{figure}

We developed an unsupervised framework, called Data Assessment Semantics (DAS), for factual validation of the  RDF statements extracted in the previous step. 
The objective is to minimize the work of domain experts during the knowledge graph construction process. Figure~\ref{fig:dasValidation} illustrates how the framework can be used to label triples as factual or non-factual.
%
%As such, DAS is domain-agnostic and can be used to validate any KG.  
The first step is to use a given RDF statement as a query in a search engine and then using the search engine's API  to retrieve the top $N$ hits containing the same keywords as in the RDF statement. The second step is to use a web page ranking tool to compute a relevance score for each retrieved page. Only the top $K$ pages with scores above a specific threshold are considered. 
The third step uses an LLM model to summarize the content of each web page and infer whether the RDF statement is factually correct based on the K top ranked web pages. The last step uses majority voting to determine whether the RDF is factual or not. The process is outlined in Algorithm~\ref{algo:dasPrompt}.
\begin{algorithm}
\caption{Evaluate the Validity of One RDF Triple.}

\begin{algorithmic}[htbp]
\State \textbf{Input:} Subject, Predicate, Object, \& Reference Text;
\State \textbf{Output:} Table with columns [Subject, Predicate, Object, Is\_The\_Triple\_Valid, Reason]
\State Initialize table with headers [Subject, Predicate, Object, Is\_The\_Triple\_Valid, Reason]
\For{\textit{each reference text}}
\State Analyze Reference Text for contextual understanding;
\If{the triple (Subject, Predicate, Object) is logically coherent in the context of Reference}
    \State Set Is\_The\_Triple\_Valid to "Yes";
    % \State Provide a detailed reason based on contextual analysis
\Else
    \State Set Is\_The\_Triple\_Valid to "No";
    % \State Provide a detailed reason based on lack of coherence or contextual support
\EndIf
\State Provide a detailed reason based on contextual analysis;
\EndFor
\State Apply  MajorityVoting Approach on Is\_The\_Triple\_Valid;
\State Add row to the table with [Subject, Predicate, Object, Is\_The\_Triple\_Valid, Reason];

\State \textbf{return} the table
\end{algorithmic}
\label{algo:dasPrompt}
\end{algorithm}

Table~\ref{tab:toolUsed} shows the  tools used to implement the DAS framework. These specific choices might be changed in future versions of ASGM-KG, as newer tools are updated or released.
With PageRank Finder, the relevance scores ranges between 0 to 10 and we set the threshold at 7 so that the most relevant page content is retrieved. 
\begin{table}
 \caption{List of software tools employed}
  \centering
  \begin{tabular}{lll}
    \toprule
    \textbf{Tool} & \textbf{Used} & \textbf{Method} \\
    \midrule
    LLM & Microsoft - Copilot PRO (GPT-4) & API \\
    Open Search Engine & DuckDuckGo & API \\
    PageRank Finder & Open Page Rank (OPR) & API \\
    \bottomrule
  \end{tabular}
  \label{tab:toolUsed}
\end{table}

\section{Validation Results} \label{sec:ExperiemtalResult}

\subsection{Validation of DAS Against Benchmark}

Using a systematic approach to  classify RDF triples as factual or non-factual is essential to our work, given the limited amount of time domain experts can dedicate to this task. As designed, the DAS framework is agnostic to a specific domain. To verify this claim, we compare its triple classification performance against other methods on CoDEx-S, a publicly available and recently published dataset derived from Wikidata and Wikipedia~\cite{codex}.
%We use CoDEx \cite{codex} to validate the performance of our proposed DAS framework. CoDEx has three knowledge graphs varying in size and structure derived from Wikidata and Wikipedia. 
CoDEx-S includes 1838 positive (factually correct) and 1838 hard negative (factually incorrect) RDF triples. Results of this comparison are shown in Table~\ref{tab:crosscheck}. The numbers in rows 1 - 5 are obtained from Table 6 in~\cite{codex}.
\begin{table}[htbp]
\centering
\caption{DAS Validation Against CoDEx-S}
\begin{tblr}{
  width = \linewidth,
  colspec = {Q[569]Q[158]Q[158]},
  cells = {c},
  hlines,
  vlines,
}
\textbf{Method}                              & \textbf{Acc} & \textbf{F1} \\
RESCAL                                     & 0.843        & 0.852       \\
TransE                                     & 0.829        & 0.837       \\
ComplEx                                    & 0.836        & 0.846       \\
ConvE                                      & 0.841        & 0.846       \\
TuckER                                     & 0.840        & 0.846       \\
{DAS - Framework\\ \textbf{(Without GT validation)}} & 0.852        & 0.836        \\
{DAS - Framework\\ \textbf{(With GT validation)}}    & 0.914        & 0.908        
\end{tblr}
\label{tab:crosscheck}
\end{table}

The without-GT-validation accuracy and F1 scores are computed by  classifying each positive and hard negative sample in CoDEx-S as factual or non-factual. 
 
They show that the performance of DAS is comparable to the benchmark methods reported in the literature.

Upon further investigation of the performance of DAS on positive triples ($\sim75\%$ true positive rate), we concluded that out of the 448 triples misclassified by DAS, 269 of them should actually be considered negative rather than positive. This was determined by manually checking each of these triples against open-source knowledge and expert opinions.
Additional details and evidence can be found in our Github repository.
Correcting the ground truth for CoDEx-S correspondingly, results in higher accuracy and F1 scores shown as with-GT-validation in Table~\ref{tab:crosscheck}.

\subsection {Validation of ASGM-KG}

As of this writing, our ASGM domain experts have been able to manually verify 579 of the 1899 DAS-validated triples in ASGM-KG as factual or non-factual.
A comparison of DAS versus domain experts on this 579 triples yields $90\%$ matching accuracy.  
The set of triples used for this comparison is admittedly small. Yet, the matching accuracy on this ground truth and on CoDEx-S are highly encouraging.
As a result, it was decided to keep all 1899 triples validated via DAS  as part of ASGM-KG.

\subsection{Downstream Tasks}
We have implemented three downstream tasks to facilitate interaction with ASGM-KG: query answering, subgraph summarization, and chat via natural language.
For query answering we use Neo4J~\cite{guia2017graph} and implemented 4 different query types:  by subject,  by object, by relation, and by subject-object. 
For subgraph summarization, we implemented $k$-hop distance summarization from a given source entity and summarization of path traversal between given source and target entities. Finally, we used Llama-3-70b-chat model for chat interaction via natural language.
Sample queries performed in ASGM-KG and subgraph summarization results can be found in our Github repository. Readers can also use our online portal~\footnote{\url{https://asgm-kg.streamlit.app/}} to interact with ASGM-KG directly via any of these downstream tasks.

\section{Discussion}

The use of LLMs for building knowledge graphs is becoming a predominant practice in natural language processing (NLP) ~\cite{li2024preliminary,kommineni2024human,Pan2023UnifyingLL}. 
For ASGM-KG, we compared the relative performances of 3 LLMs and chose GPT 4 due to its contextual understanding and high performance on ambiguous text. This choice is likely to change as newer LLM technology becomes available. 
Similarly, the choice of specific tools for factual reduction is not fixed. 

The DAS framework was created as a means to help reduce the workload of ASGM domain experts. It systematizes a standard process that  many people use to determine veracity of a natural language statement. Others have proposed and are using similar approaches, see e.g.~\cite{kg-validator}. It is entirely possible that for a given statement, DAS may output no classification at all, e.g., if a web search yields no results. It is also possible that DAS may output an erroneous classification. Concerning cases include the unintended use of web misinformation. But it is unreasonable to expect a domain expert to factually verify by hand the large number of statements that can be generated by an LLM, or that any one expert, no matter how well versed in the material, is omniscient.

The resource presented in this paper is currently hosted online and implemented via standard tools. It performs several down-stream tasks including QnA, graph summarization, and natural language interaction. 
ASGM is a large and ever-growing global environmental and social issue, particularly in the global tropics [e.g.~\cite{dethier2023global}], and we expect stakeholders in Madre de Dios and other areas around the globe where ASGM is prevalent to explore our knowledge base and extract useful information, and to contribute to its growth. How this resource influences future policy is an important topic of interest. Our CINCIA partners are committed to maintaining this resource and collecting use statistics. We are also committed to  increasing the size of ASGM-KG with the inclusion of additional documents curated by domain experts worlwide. 
\section{Conclusion}
In this paper, we have introduced a novel resource, ASGM-KG, an application of knowledge graphs to help understand the growing threat to the environment and to human populations by artisanal and small-scale gold mining. 
ASGM-KG was constructed in collaboration with ASGM domain experts. It is currently deployed as a freely available web application implementing three important downstream tasks: QnA, graph summarization, and natural language dialog.

Expanding ASGM-KG is an important objective to support both stakeholders as well as the research community.  New approaches, such as expert-in-loop~\cite{Rahman2024KnowledgeAA, Manzoor2022ExpandingKG}, may be considered to help domain experts curate the large amounts of information that can be extracted and summarized via tools such as the DAS framework or inferred via link prediction~\cite{Pahuja,Kazemi}. 

At the same time, there is still a great need for intelligent methods that can help accurately forecast future land cover change, loss of habitat, and water quality resulting from human activity and natural phenomena. These approaches should be capable of dealing with and extracting information from the large amounts of complex and uncertain data available for the study of ASGM, such as  multimodal, multi-resolution, time- and geolocation-dependent image data, in addition to text data from reports, news media, interviews, etc. Neuro-symbolic methods can be explored to effectively integrate such data types for improved interpretability, explainability, and contextual awareness.

\bibliographystyle{unsrt}  
\bibliography{ref}

\end{document}